\documentclass[11pt,a4paper]{article}
\usepackage{times,latexsym}
\usepackage{url}
\usepackage[T1]{fontenc}

   \usepackage[acceptedWithA]{tacl2021v1}
\usepackage{tacl2021v1}
\usepackage{xspace,mfirstuc,tabulary}

\newif\iftaclinstructions
\taclinstructionsfalse % AUTHORS: do NOT set this to true
\iftaclinstructions
\renewcommand{\confidential}{}
\renewcommand{\anonsubtext}{(No author info supplied here, for consistency with
TACL-submission anonymization requirements)}
\newcommand{\instr}
\fi

\iftaclpubformat % this "if" is set by the choice of options

\else

\fi

\usepackage[utf8]{inputenc}

\usepackage{microtype}

\usepackage{graphicx}
\usepackage{xcolor}
\usepackage{amsfonts}
\usepackage{mathrsfs}
\usepackage{float}
\usepackage{paralist}
\usepackage{caption}
\usepackage{subcaption}
\restylefloat{table}
\usepackage{listings}% http://ctan.org/pkg/listings
\newcommand*{\ttfamilywithbold}{\fontfamily{lmtt}\selectfont}
\usepackage[arabic,russian,british,finnish]{babel}
\babelprovide[import,main]{bengali}
\babelprovide[import,main]{arabic}
\babelprovide[import]{hindi}
\babeltags{english=english}
\babeltags{bengali=bengali}

\title{\textsc{QAmeleon} \includegraphics[height=2\fontcharht\font`\B]{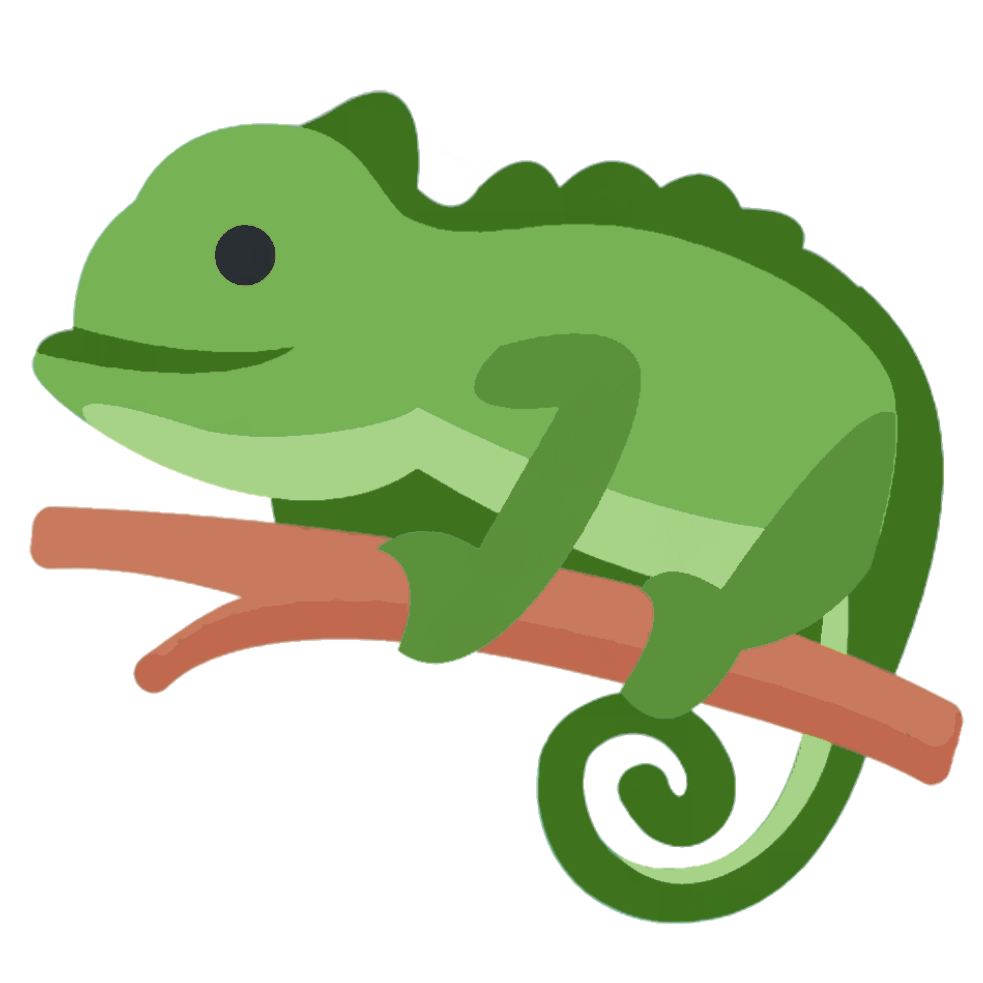}:
Multilingual QA  with Only 5 Examples}

\definecolor{forestgreen}{HTML}{009B55}
\usepackage{todonotes}

\newcommand{\tydiqa}{\textsc{TyDi QA}}
\newcommand{\tydiqagoldp}{\textsc{TyDiQA-GoldP}}

\author{Priyanka Agrawal\thanks{\ \ Equal contribution. See Contributions section for details.} \ \ \ \ 
        Chris Alberti$^{*}$ \ \ \ 
        Fantine Huot \ \ \ \
        Joshua Maynez \\
        {\bf Ji Ma} \ \ \ \  
        {\bf Sebastian Ruder} \ \ \ \ 
        {\bf Kuzman Ganchev} \ \ \ \ 
        {\bf Dipanjan Das} \ \ \ \ 
        {\bf Mirella Lapata} \\
  Google Research\\
  \texttt{\normalsize \{priyankagr,chrisalberti,fantinehuot,joshuahm,} \\
  \texttt{\normalsize maji,ruder,kuzman,dipanjand,lapata\}@google.com}
  }

\date{}

\begin{document}
\maketitle
\begin{abstract}

The availability of large, high-quality datasets has been a major driver of recent progress in question answering (QA).
Such annotated datasets, however, are difficult and costly to collect,
and rarely exist in languages other than English, rendering QA
technology inaccessible to underrepresented languages. An alternative
to building large monolingual training datasets is to leverage
pre-trained language models (PLMs) under a few-shot learning setting. Our approach, \textsc{QAmeleon}, uses a PLM to automatically \emph{generate} multilingual data upon which QA
models are fine-tuned, thus avoiding costly annotation. Prompt tuning
the PLM  with only five examples per language delivers accuracy
superior to translation-based baselines; it  bridges nearly~60\% of the gap between
an English-only baseline and a fully-supervised upper bound fine-tuned on  almost~50,000 hand-labeled examples; and consistently leads to  improvements compared to directly fine-tuning a QA model on labeled examples in low resource settings. Experiments on the \tydiqagoldp{} and
\textsc{MLQA} benchmarks show that few-shot prompt tuning for data synthesis scales across
languages and is a viable alternative to large-scale annotation.\footnote{We release the multilingual QA synthetic data used for fine-tuning them at \url{https://github.com/google-research-datasets/QAmeleon}.}

\end{abstract}

\section{Introduction}

Question answering (QA) has seen impressive progress in recent years
enabled by the use of large pre-trained language models
\cite{devlin-etal-2019-bert,lewis-etal-2020-bart,Raffel:ea:2020}, and
the availability of high-quality benchmarks
\cite{rajpurkar-etal-2016-squad,trischler-etal-2017-newsqa,kwiatkowski-etal-2019-natural}. Many
QA datasets frame the task as reading comprehension where the question
is about a paragraph or document and the answer is a span therein.
Advances in QA modeling have been primarily reported for English,
which offers a considerable amount of high-quality training data
compared to other languages.  More recently, efforts have focused on
the creation of \emph{multilingual} QA benchmarks such as \tydiqa{} (10~languages; \citealt{clark-etal-2020-tydi}), \textsc{MLQA}
(6~languages; \citealt{lewis-etal-2020-mlqa}), and \textsc{XQuAD}
(10~languages; \citealt{artetxe-etal-2020-cross}). Among these, only \tydiqa{} is genuinely large-scale, \textsc{MLQA} and \textsc{XQuAD} are
limited to an evaluation set due to the high cost and labor required
to collect data across languages.

As a result, efforts to localize QA models to new languages have been
primarily focusing on \emph{zero-shot} approaches. Recent proposals
include using machine translation to approximate training data for
supervised learning \cite{lewis-etal-2020-mlqa}, and data augmentation
via generating synthetic questions for new languages
\cite{riabi-etal-2021-synthetic,shakeri-etal-2021-towards}. Both
approaches rely on transfer from English, which leads to a dependence on translation 
artifacts \citep{koppel-ordan-2011-translationese,artetxe-etal-2020-translation} and a bias towards the linguistic characteristics of English, which is not the
best source for all target languages \cite{Lin2019}. However,
annotating a minimally-sized data sample can potentially overcome
these limitations while incurring significantly reduced costs compared
to full dataset translation \citep{garrette-baldridge-2013-learning}.

\begin{figure*}[ht]
\centering
\includegraphics[width=\textwidth]{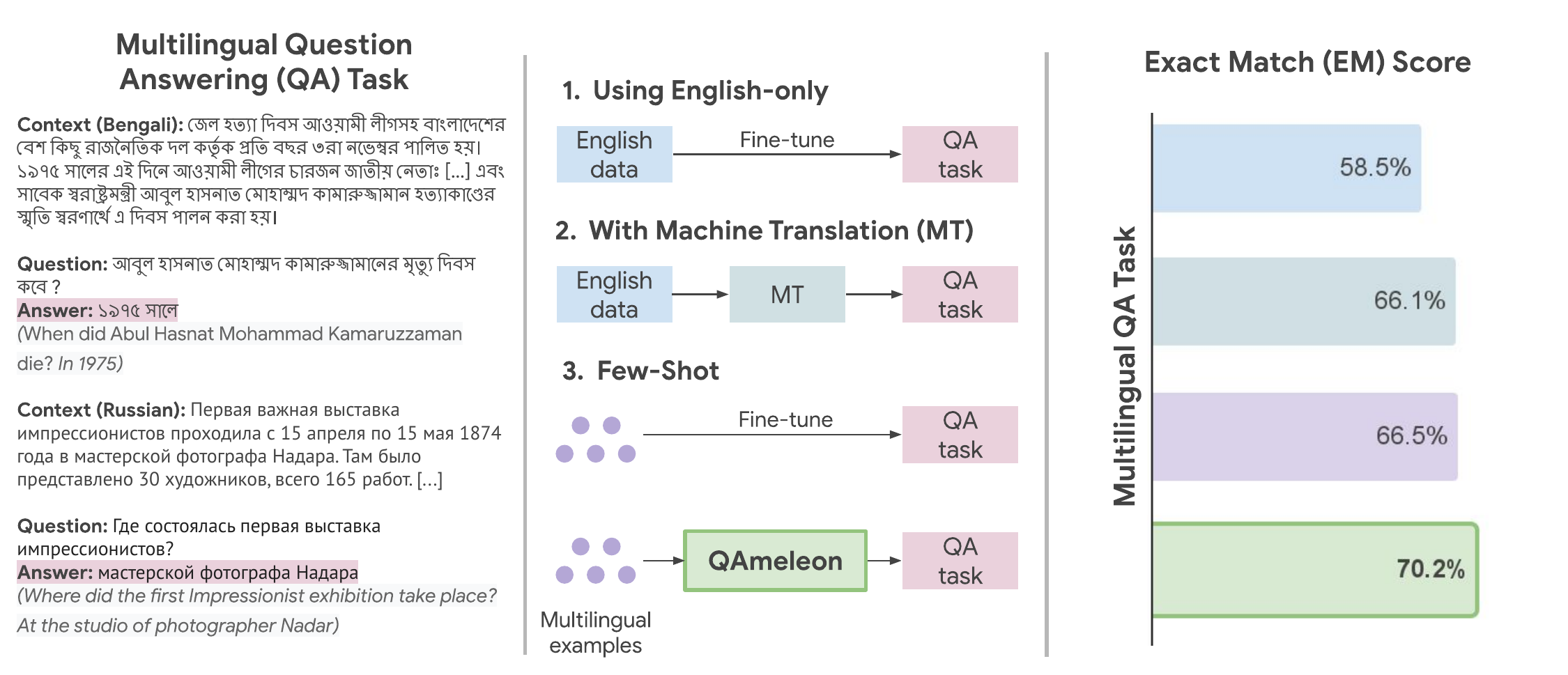}
\vspace{-0.5cm}
\caption{Synthetic data generation for multilingual question-answering
  (QA). Left: Examples of the multilingual QA task. Translations are
  added for readability. Middle: Strategies for localizing QA models
  to new languages: 1. Using English QA data as a zero-shot approach,
  2. with Machine Translation (MT) to approximate training data for
  supervised learning, and 3. few-shot approaches with a handful of
  multilingual examples. Right: Model performance on the multilingual QA
  task. We report average Exact Match (EM)  across all
  languages on the \tydiqagoldp{} dataset \cite{clark-etal-2020-tydi}.  }
\label{fig:approach}
\end{figure*}

In this paper, we argue that a few-shot approach in combination with
synthetic data generation and existing high-quality English resources
can mitigate some of the above mentioned artifacts. Beyond question
answering, multilingual approaches have succeeded at leveraging a
small number of annotations within a variety of tasks
\citep[\emph{inter alia}]{zhao-etal-2021-closer} including natural
language inference, paraphrase identification, and semantic parsing
\cite{Sherborne:Lapata:2022}. Existing work \citep[\emph{inter
    alia}]{brown2020language,schick-schutze-2021-just} has further
shown that prompting pre-trained large language models (PLMs) can lead
to strong performance on various tasks, including question answering
\cite{khashabi-etal-2020-unifiedqa,chowdhery2022palm} and open-ended
natural language generation
\cite{Tang:ea:2022,Yang:ea:2022}. Investigations of prompting in
multilingual settings have also shown strong few-shot performance in
classification tasks \cite{winata-etal-2021-language}, natural
language inference \cite{zhao-schutze-2021-discrete}, common sense
reasoning \cite{Shi:ea:2022}, machine translation \cite{Lin2022},
and retrieval \cite{dai2022promptagator}.

We synthesize these directions into \textsc{QAmeleon}, an approach for
bootstrapping multilingual QA systems, with as few as five examples in
a new target language (see Figure~\ref{fig:approach}). We use gold
annotations to prompt-tune a PLM in order to automatically generate
multilingual QA data, which is then used to fine-tune a QA model. We find
that \textsc{QAmeleon} delivers accuracy superior to zero-shot methods
and competitive translation-based baselines, and in some cases
competes with the fully supervised upper bound.\footnote{This is noteworthy as multilingual models fine-tuned on translated data---also known as translate-train---form the state of the art on most multilingual datasets \cite{ruder-etal-2021-xtreme}.}
Experiments on the
\tydiqa{} \cite{clark-etal-2020-tydi} and \textsc{MLQA}
\cite{lewis-etal-2020-mlqa} benchmarks show that few-shot prompt
tuning \cite{lester-etal-2021-power} scales across languages,
significantly outperforms prompt engineering \cite{brown2020language}
with the same number of labeled examples, and is a viable alternative
to large-scale annotation.

Our contributions include (a)~a new approach to
bootstrapping a multilingual QA system; \textsc{QAmeleon} 
prompt-tunes a PLM with as few as five gold
examples to automatically generate multilingual QA data which is then
used to fine-tune a QA model; (b)~a series of experimental results showing
significant improvements over existing approaches in the few-shot
regime, ranging from 12\% absolute accuracy on \tydiqagoldp{}
\cite{clark-etal-2020-tydi} over an English-only baseline and 4\% absolute
accuracy over a competitive translate-train baseline; (c)~extensive analysis
of the behavior of \textsc{QAmeleon} in zero shot
and low resource regimes, on different multilingual QA datasets, and
in comparison to prompt-engineering. 

\section{Synthetic Data Generation}
\label{sec:method}

Let~$\mathcal{D}_{l}$ denote a QA dataset with examples provided by
human annotators, where $l$ is a \emph{target} language in a set~$L$
of languages of interest. $\mathcal{D}_{l}$ consists of samples~$(c,
q, a)_l$, where~$c$ is a paragraph of text, $q$~is a question, and
$a$~is an answer extracted from~$c$ (see Figure~\ref{fig:approach}
left). We further use~$\mathcal{D}_{l,n}$ to denote a
dataset $\mathcal{D}_{l}$, but making $n$ explicit, with~$n$ referring to the number of
examples it contains. For instance, $\mathcal{D}_{\mathrm{fr},5}$ 
denotes a French QA dataset with 5 examples. Finally,
let~$\mathcal{U}_{l}$ denote sets of \emph{unlabeled} paragraphs in
language $l$; we assume these are in-domain with respect to the
paragraphs in~$\mathcal{D}_{l}$ but are not accompanied by questions
or answers.

Throughout this work, we will assume the availability
of~$\mathcal{D}_{\mathrm{en}}$, a large QA dataset in English
(\emph{source} language). This assumption corresponds to the
observation that most large-scale QA datasets
\cite{rajpurkar-etal-2016-squad,yang-etal-2018-hotpotqa,bajaj2016ms,kwiatkowski-etal-2019-natural} contain examples exclusively in
English.
%The dataset contains examples of the form $(c, q, a)_{en}$, where $c$~is a text passage, $q$~is a question, and $a$ is an answer extracted from~$c$
For languages other than English, we assume that only small datasets~$\mathcal{D}_{l,n}$ are available for training (e.g., $n=5$)
(``Few-Shot'' scenario) or no data at all (``English-Only'' scenario).
We will also assume that sets~$\mathcal{U}_{l}$ of
unlabeled passages are available for all target languages.
Our task will be to synthesize QA data in each \emph{target}
language~$l$ in order to fine-tune QA models on~$l$ directly.

In the rest of this section we formally describe three ways of synthesizing QA data and
give further details on the two scenarios we consider, ``English-Only'' and ``Few-Shot''.

\label{sec:synth-data-gener}

\subsection{Machine Translation (MT)} A widely adopted approach
\cite{lewis-etal-2020-mlqa,shakeri-etal-2020-end} makes use of a
machine translation system~$\mathcal{T}$ to automatically translate
text from one language into another.
Let~$\mathcal{T}_{l'}(\mathcal{D}_l)$ denote the translation of
dataset $\mathcal{D}_l$ from language~$l$ to language~$l'$. The
translation is performed by independently applying $\mathcal{T}$ to
context $c$, question~$q$, and answer~$a$ for each example in the
source dataset (see approach 2 in Figure~\ref{fig:approach}). A
synthetic QA dataset~$\mathcal{D}_\mathrm{MT}$ is generated by
translating the entire English dataset into each language of interest:
\[
\mathcal{D}_\mathrm{MT} = \mathcal{D}_\mathrm{en} \cup \bigcup_{l \in
  L - \{\mathrm{en}\}} \mathcal{T}_{l}(\mathcal{D}_\mathrm{en}).
\]

The approach described here is known as ``translate-train''. An
alternative is ``translate-test'', where translation is employed
during inference instead of training. Multilingual inputs are
translated to English and inference is done via an English QA
model. The English predictions are then translated back to the
respective target language. We experimentally found ``translate-test''
to perform poorly on our task in comparison to translate-train due to
its reliance on multiple noisy translation steps.

Note that fine-tuning on~$\mathcal{D}_\mathrm{MT}$ still relies on the
support of the high-quality~$\mathcal{D}_\mathrm{en}$.  Previous work
\cite{kramchaninova-defauw-2022-synthetic, vu-2022-zeroshot} has
highlighted various limitations with multilingual approaches based on
MT including (a)~their dependence on the quality of available MT
systems in a given language and in turn the availability of
high-quality (expensive) parallel data, (b)~a potential misalignment
of answer spans after translation in context to the passage vs
translation of answers independently, and (c)~translationese artifacts
and English-centric content topics \cite{clark-etal-2020-tydi}.

\subsection{Prompt Engineering (PE)}

PLMs \cite{brown2020language,chowdhery2022palm} have recently shown
unprecedented performance on a vast number of tasks, including natural
language generation, without the need for modifying any of the model's
parameters, simply by hand-designing a textual prompt that
instructs the model to perform a certain task. Following
\newcite{brown2020language}, we consider a class of hand-designed
prompts referred to as  ``prompting'' or  ``in-context learning''. The prompt starts with
a free form instruction, followed by a small number of instances
exemplifying how the task is solved. An incomplete instance is then
appended to this prompt and the PLM performs the task by completing
that instance. We refer to this approach as ''prompt engineering''
(PE), since the input to the PLM has to be hand-engineered based on
human intuition about the target task (see approach 3 in
Figure~\ref{fig:approach}).

In order to hand-engineer prompts for our task, we use a small set of
parallel examples $\mathcal{C}_{l,n}$ consisting of passages, questions, and their answers
in the English source and target language~$l$. We discuss how we
construct these examples shortly. For now, suffice it to say that we
create two prompts for answer and question generation,
respectively.\footnote{We find that joint answer and question generation using single-stage prompting performs worse in comparison to two-stage generation.} 
Our first prompt is used to obtain an answer~$a_l$ in
the target language~$l$ from passage~$c_l$:
\begin{lstlisting}
I will write potential answers
for the following passages.
  Passage: $c_l$
  Answer in English: $a_\mathrm{en}$
  Answer in the original language: $a_l$
...
\end{lstlisting}
The second prompt generates question~$q_l$, utilizing passage~$c_l$
and the previously predicted answer~$a_l$:

\begin{lstlisting}
I will write questions and answers
for the following passages.
  Passage: $c_l$
  Answer: $a_l$
  Question in English: $q_\mathrm{en}$
  Question in the original language: $q_l$
...
\end{lstlisting}
We generate synthetic data instances $(c,q,a)_l$ where~$a$ and~$q$ are
inferred by applying our two prompts consecutively on each
passage~\mbox{$c_l \in \mathcal{U}_{l}$} (recall $\mathcal{U}_{l}$ is
the set of unlabeled passages in target language~$l$). 

In the English-Only scenario, neither questions nor answers are available in target language; we 
obtain these by resorting to machine translation:
\begin{eqnarray*}
  & & \mathcal{C}^\mathrm{en-only}_{l,n} = \\
  & & \{(\mathcal{T}_l(c), q, a, \mathcal{T}_l(q), \mathcal{T}_l(a)) | (c, q, a) \in \mathcal{D}_{\mathrm{en}, n}\},
\end{eqnarray*}
In the ``Few-Shot'' setting, we have access to \mbox{$n$-labeled} examples (questions and answers) in the
target language, and translate these into English: 
\begin{eqnarray*}
  & & \mathcal{C}^{n\mathrm{-shot}}_{l,n} = \\
  & & \{(c, \mathcal{T}_\mathrm{en}(q), \mathcal{T}_\mathrm{en}(a), q, a) | (c, q, a) \in \mathcal{D}_{l, n}\}.
\end{eqnarray*}
Let~$\mathcal{P}^e_l$ denote this prompting based generation. We can write the
generated synthetic dataset as: 
\[\mathcal{D}_\mathrm{PE} = \mathcal{D}_\mathrm{en} \cup \bigcup_{l \in L - \{\mathrm{en}\}} \mathcal{P}^e_l(\mathcal{U}_l).\]

Note that in the composition of the prompt, we always include English as an intermediate or ``bridge'', i.e., asking the model to predict questions and answers in English in addition to the ones in the target language, as we experimentally found it improves the quality of the generated data.  The use of a bridge for this task can be thought of as an
example of multilingual ``chain of thought'' prompting
\cite{wei2022chain}.

\subsection{\textsc{QAmeleon} (PT)}

In this approach, an optimizer is utilized
to minimize the cross-entropy loss by updating the PLM's parameters for $P(a, q | c, l)$
over a training set containing examples $(c,q,a)_l$ for the languages in
$L$. As with PE, we generate the training set for the PLM in 
two ways. For ``English-Only'' we construct the dataset as $\bigcup_{l
  \in L} \mathcal{T}(\mathcal{D}_\mathrm{en})$, while for ``Few-Shot''
we use $\bigcup_{l \in L} \mathcal{D}_{l,n}$.

Given the small size of
the training set in the ``Few-Shot'' setting and the large size of current models, we opt for using prompt
tuning \cite[PT;][]{lester-etal-2021-power}, a parameter-efficient fine-tuning
variant where we concatenate a soft prompt of length $m$
tokens to the input of the PLM, where $m$ is a hyperparameter always set to 50 in this work. Only the embeddings of these $m$ prompt tokens are allowed to be modified by the
optimizer.
We note that in prompt tuning, like in prompt engineering, the parameters of the PLM remain unchanged. What is trained is only the short soft prompt that is prepended to the input embeddings at inference time.

We use~$\mathcal{P}^t_l$ to denote the operation of generating question-answer pairs through greedy decoding on the prompt-tuned PLM, by taking an unlabeled
passage $c_l \in \mathcal{U}_{l}$ as input, preceded by a few tokens
encoding language~$l$. We finally obtain the synthetic QA
dataset~$\mathcal{D}_\mathrm{PT}$ as: 
\[\mathcal{D}_\mathrm{PT} = \mathcal{D}_\mathrm{en} \cup \bigcup_{l \in L - \{\mathrm{en}\}} \mathcal{P}^t_l(\mathcal{U}_l).\]

\subsection{Data Assumptions}
\label{sec:assumptions}

\paragraph{English-Only} In this scenario, only training data in
    English is available, denoted as $\mathcal{D}_\mathrm{en}$. Prompt
    Engineering (PE) assumes parallel exemplars are available, while
    Prompt Tuning (PT) requires exemplars in the target language
    only. Both are possible by
    translating examples of the English data $\mathcal{D}_\mathrm{en}$ into each target
    language. Machine
    Translation (MT) approaches in this work  follow this scenario only.

\paragraph{Few-Shot}  In this scenario, a small number of examples ($n$-shot) are available in each target language, denoted as
  $\mathcal{D}_{\mathrm{l},n}$.  In this scenario, parallel exemplars
  for Prompt Engineering (PE) can be obtained by translating the
  target language data into English.
  Prompt Tuning (PT) only requires exemplars in the target language, which are readily available in this setting.

\section{Experimental Setup}
\label{sec:exp}

We evaluate the synthetic data generation approaches presented in
Section~\ref{sec:synth-data-gener} across various languages on two
benchmark datasets, which we discuss below. We also describe various
model configurations, and comparison systems before presenting our
results.

\subsection{Datasets}
\label{sec:datasets}

\begin{table}[t]
    \centering \resizebox{\columnwidth}{!}{
    \begin{tabular}{l|cc|cc} 
                %  & TyDi-QA & TyDi-QA & MLQA & MLQA \\
 & \multicolumn{2}{c|}{\textbf{\tydiqagoldp{}}} & \multicolumn{2}{c}{\textbf{\textsc{MLQA}}} \\
       \multicolumn{1}{c|}{\textbf{Language}}  & \multicolumn{1}{c}{Train}   & \multicolumn{1}{c}{Eval}     & \multicolumn{1}{|c}{Dev} & \multicolumn{1}{c}{Test} \\
       \hline \hline
         Arabic & 14,805 & 921 & 517 & 5,335 \\
         Bengali & 2,390 & 113 & --- & --- \\
         Chinese & ---& ---& 504 & 5,137 \\
         English & 3,696 & 440 & 1,148 & 11,590 \\
         Finnish & 6,855 & 782 & --- & --- \\
         German & --- & --- & 512 & 4,517\\
         Hindi & ---& --- & 507 & 4,918 \\
         Indonesian & 5,702 & 565 & --- & --- \\
         Kiswahili & 2,755 & 499 & --- & ---\\
         Korean & 1,625 & 276 & --- & ---\\
         Russian & 6,490 & 812 & --- & ---\\
         Spanish & --- & --- & 500 & 5,253  \\
         Telugu & 5,563 & 669 & --- & --- \\
         Vietnamese & --- & --- & 511 & 5,495 \\
       \hline
         Total & 49,881 & 5,077 & 4,199 & 42,245 \\ \hline \hline
    \end{tabular}}
    \caption{Number of question-answer pairs per language and data split for the datasets considered in this work.}
    \label{tab:data}
\end{table}

\paragraph{\tydiqa{}} \hspace*{-.25cm}\cite{clark-etal-2020-tydi}
is a multilingual extractive question answering dataset designed to
represent a typologically diverse set of languages. Annotators were
given a Wikipedia passage in the target language and asked to write a
question that could not be answered by that passage. For each question, the top-ranked Wikipedia article was then retrieved via Google Search.
Annotators were subsequently asked to answer the question given the
retrieved Wikipedia article. As a result of this information-seeking task design,
questions in \tydiqa{} are often without an answer. In this work
we consider \tydiqagoldp{}: the Gold Passage version of \tydiqa{} where only
questions with answers in the Wikipedia page are given and the model
has to identify the answer in the passage that contains
it (see Table~\ref{tab:data} for statistics on this dataset).

\paragraph{\textsc{MLQA}} \hspace*{-.25cm}\cite{lewis-etal-2020-mlqa}
is an extractive question answering dataset, designed for
evaluating multilingual and cross-lingual question answering
models. \textsc{MLQA} does not publish a training split, but only
development and test partitions. \textsc{MLQA} was created by aligning
sentences in Wikipedia passages across different
languages. Annotators then created questions based on English
sentences, professional translators translated these questions to
other languages, and finally annotators selected answers from passages
containing sentences aligned to the translated questions. As in
\tydiqagoldp{}, the task is to extract the answer from a passage given a question
(dataset statistics are shown in Table~\ref{tab:data}).

\paragraph{Unlabeled Data} We obtained paragraphs~$\mathcal{U}_l$ in each
target language from Wikipedia. Specifically, we pre-processed
Wikipedia pages using WikiExtractor \cite{Wikiextractor2015}. 
Paragraphs were
sampled uniformly, with a length between~200 and 510~characters. The
target language was determined based on the language code of the
Wikipedia edition.

\begin{table*}[t]
    \centering
    \begin{tabular}{l|ccc|ccc} 
    \multicolumn{1}{c|}{} & \multicolumn{3}{c|}{\textbf{English-Only}} & \multicolumn{3}{c}{\textbf{Few-Shot}} \\
       \multicolumn{1}{c|}{\textbf{Method}}     & Translate & Avg EM &  Avg F1 & n-Shot & Avg EM &  Avg F1\\
       \hline \hline
        % Rows below are generated by a script.
        %%%%%%%%%%%%%%%%%%
		Baseline  &    & $58.5_{(\pm 3.1)}$ & $74.2_{(\pm 2.6)}$ & 5 & $66.5_{(\pm 0.7)}$ & $79.8_{(\pm 0.4)}$ \\
		MT  &  \checkmark  & $66.1_{(\pm 2.1)}$ & $79.5_{(\pm 1.8)}$ & 5 & --- & --- \\
		PE  &  \checkmark  & $64.4_{(\pm 1.4)}$ & $76.9_{(\pm 1.1)}$ & 5 & $62.6_{(\pm 1.8)}$ & $77.6_{(\pm 1.2)}$ \\
		PE + MT  &  \checkmark  & $\textbf{69.4}_{(\pm 0.4)}$ & $\textbf{81.4}_{(\pm 0.4)}$ & 5 & $67.9_{(\pm 0.2)}$ & $80.5_{(\pm 0.6)}$ \\
		\textsc{QAmeleon} (PT)  &  \checkmark  & $65.5_{(\pm 0.7)}$ & $79.4_{(\pm 0.7)}$ & 5 & $70.2_{(\pm 0.2)}$ & $81.7_{(\pm 0.1)}$ \\
		\textsc{QAmeleon} (PT)+MT  &  \checkmark  & $68.1_{(\pm 0.8)}$ & $80.9_{(\pm 0.7)}$ & 5 & $\textbf{70.7}_{(\pm 0.9)}$ & $\textbf{82.2}_{(\pm 0.8)}$ \\
		%%%%%%%%%%%%%%%%%%
       \hline
        code-davinci-002$\S$ &     & --- & ---  &   1 &   48.1 &     --- \\
        PaLM-540B$\dag$ &     & --- & ---  &   1--10 &   60.0 &     --- \\
        Flan-U-PaLM-540B$\ddag$ &     & --- & ---  &   1 &   68.3 &     ---
       \\\hline \hline
    \end{tabular}
    \caption{Synthetic question-answering data generation methods for
      training multilingual reading comprehension systems on
      \tydiqagoldp{}. We report averages over 3 runs of
      fine-tuning mT5-XL on gold or synthetic data. Standard deviation is given in parentheses.  Performance for individual languages (excluding English) is shown in Table~\ref{tab:lang}.  For comparison we also include recent few-shot prompting results with large language models on \tydiqagoldp{}: \newcite{chen2021evaluating}$\S$, \newcite{chowdhery2022palm}$\dag$, and \newcite{chung2022scaling}$\ddag$.
      }
    \label{tab:main_eval}
\end{table*}

\subsection{Model Configuration}
\label{sec:model-configuration}

\paragraph{Synthetic Data Generation} 
 In our \tydiqa{} experiments, we treat the English training
 data as the English source. For \textsc{MLQA}, we employ the English
 \textsc{SQuaD} \cite{rajpurkar-etal-2016-squad} training data as the source.  In
 the Few-Shot scenario, our human-annotated target-language examples
 $\mathcal{D}_{l,n}$ are taken from the training split of
 \tydiqagoldp{} and the validation split of \textsc{MLQA}.

For machine translation (MT), we employ the public Google Translate
API \cite{wu2016google} while the PLM utilized in this work is PaLM-540B \cite{chowdhery2022palm}. We perform heuristic checks to
clean synthetic datasets~$\mathcal{D}_\mathrm{PE}$
and~$\mathcal{D}_\mathrm{PT}$.  We only preserve a question-answer
pair if the generated answer~$a$ is a substring of the given
context~$c$, but not a substring of the query~$q$. We perform the
first check as both \tydiqagoldp{} and \textsc{MLQA}
are extractive QA datasets. We perform the latter check because we
empirically found that some of the low quality generated
question-answer pairs were trivially answered based on the content of
the question alone, for example, $q$: ``where is X?'', $a$:~``X''.

In the construction of $\mathcal{D}_\mathrm{PE}$, we additionally
perform round-trip filtering \cite{alberti-etal-2019-synthetic} as 
qualitative analysis of random QA pairs suggested a higher level of
noise in the PE-generated data. This round-trip consistency check is
done by comparing the originally generated answer $a$ in $(c,q,a)_l$ with
the predicted answer. This predicted answer is obtained by prompting
the PLM to answer question $q$ based on passage $c$. We also tried
round-trip filtering for PT generated data,
however, we did not observe any gains. We report detailed
statistics of the synthetically generated datasets in
Section~\ref{sec:data-analysis}.

In the construction of $\mathcal{D}_\mathrm{PT}$, we prompt-tune the
PLM on $\bigcup_{l \in L} \mathcal{T}(\mathcal{D}_\mathrm{en})$ or
$\bigcup_{l \in L} \mathcal{D}_{l,n}$ as detailed earlier. Prompt
tuning is performed with the AdaFactor optimizer
\cite{shazeer2018adafactor}. We tune a prompt of length $m=50$ tokens for
up to 1,000 steps, evaluating every 50 steps, with a batch size of 16
examples, and learning rate of 0.3 with a linear warmup of 200
steps. We use early stopping to select the best prompt per language
based on BLEU \cite{papineni2002bleu} on a held-out dataset from the English \tydiqagoldp{}, translated to each target language.

\paragraph{Question Answering}
 We fine-tuned an mT5-XL model \cite{xue2021mt5} for question-answering
 to evaluate different synthetic data generation methods
 ($\mathcal{D}_\mathrm{MT}$, $\mathcal{D}_\mathrm{PE}$, and
 $\mathcal{D}_\mathrm{PT}$). As a baseline, we further use mT5-XL fine-tuned on available training data. Specifically, in the English-Only scenario, Baseline mT5-XL is
fine-tuned on the English QA data $\mathcal{D}_{en}$. In the Few-shot
scenario, Baseline mT5-XL is fine-tuned on $n$~human annotated examples in the
target languages (same number given to PE and PT). We conducted experiments on
 \tydiqagoldp{} \cite{clark-etal-2020-tydi} and \textsc{MLQA}
 \cite{lewis-etal-2020-mlqa}, see
 Section~\ref{sec:datasets}.

Throughout downstream QA evaluation, mT5-XL was fine-tuned with AdaFactor, with a learning rate of 0.0002, a batch size of 64, up to 3,000 and 5,000 steps of training for \tydiqagoldp{} and \textsc{MLQA} respectively evaluating every 50 steps. We measure QA performance with Exact Match (EM) and F1, and report the unweighted average across languages (excluding English). For \tydiqagoldp{}, we report results on the development split which is commonly used as an evaluation set since the test split is unavailable. We select mT5 checkpoints per language using EM, and report the average of 3~runs.  For MLQA, we present results on the test split, selecting the best mT5 checkpoint based on the average EM on the MLQA dev set.

\begin{table*}[t]
    \centering
    \begin{tabular}{l|ccccccccc|c} 
       \textbf{Method}  & \textbf{n-shot}   &  \textbf{Ar} & \textbf{Bn} & \textbf{Fi} & \textbf{Id} & \textbf{Ko} & \textbf{Ru} &  \textbf{Sw} & \textbf{Te} & \textbf{Avg} \\
       \hline \hline
       % Rows below are generated by a script.
       %%%%%%%%%%%%%%%%%%%%%%%%%%%%%
		Baseline & 5 & 65.9 & 68.4 & 65.1 & 71.3 & 68.4 & 57.6 & 60.1 & 75.4 & 66.5 \\
		MT & 0 & 66.3 & 62.2 & 65.2 & 72.4 & 63.9 & 61.1 & 70.5 & 67.0 & 66.1 \\
		PE & 0 & 60.4 & 66.7 & 63.5 & 63.6 & 65.1 & 53.8 & 74.5 & 67.3 & 64.4 \\
		PE + MT & 0 & \textbf{68.1} & 70.5 & 68.2 & 73.6 & \textbf{68.5} & 61.0 & \textbf{78.4} & 66.9 & 69.4 \\
		\textsc{QAmeleon} (PT) & 5 & 65.4 & \textbf{76.7} & \textbf{69.4} & 69.0 & 67.6 & 61.5 & 75.6 & \textbf{76.7} & 70.2 \\
		\textsc{QAmeleon} (PT)+MT & 5 & 67.9 & 72.6 & 69.2 & \textbf{73.8} & 65.1 & \textbf{62.8} & 77.7 & 76.1 & \textbf{70.7} \\
       %%%%%%%%%%%%%%%%%%%%%%%%%%%%%
    %   Baseline & 5 &  62.6 & 63.7 & 61.4 & 66.7 & 66.7 & 56.0 & 56.9 & 71.3 & 63.7 \\
    %   MT    & 0                 &  \textbf{66.5} & 64.3 & 65.0 & \textbf{72.3} & 63.5 & {60.4} & 72.6 & 63.8 & 66.0  \\
    %   PE   & 0                     & 62.7 & 67.6 & 64.3& 66.4 & \textbf{67.9 }& 52.6 & {74.0} & 66.2 & 65.2 \\
    %   \textsc{QAmeleon} (PT) & 5 & 66.2 &\textbf{ 77.3 }& 69.1 & 69.9 &67.8 & 61.9 & 75.9 & \textbf{76.9} & 70.6 \\
    %   \textsc{QAmeleon} (PT) + MT & 5& 65.9	& 75.5 & \textbf{70.9}&	{71.0} &	67.1 &\textbf{	62.0 }&	\textbf{76.0 }&	\textbf{76.9} & \textbf{70.7 }\\
       %%%%%%%%%%%%%%%%%%%%%%%%%%%%%
       \hline
       Supervised & Multi-k & 75.7 & 81.4 & 74.5 & 79.8 & 77.2 &72.8 &
       82.6 & 83.0 & 78.4 \\
       \hline
       \% tokens in PLM & --- & \textit{0.15} &\textit{ 0.03} & \textit{0.42} & \textit{0.16} & \textit{0.19} & \textit{0.53} & \textit{0.01} & \textit{0.02} & --- 
       \\ \hline \hline
    \end{tabular}
    \caption{QA performance (Average EM over three runs) for individual languages on the
      \tydiqagoldp{} evaluation set; the backbone of the QA model is 
      mT5-XL  fine-tuned on gold (Baseline, Supervised) or synthetically
      generated data. The final row displays the percent of tokens for
      each language in the PLM training data.
    }
    \label{tab:lang}
\end{table*}

\section{Results}

\paragraph{\textsc{QAmeleon} (PT) Delivers the Best QA
  System} Table~\ref{tab:main_eval} summarizes our results on
\tydiqa{} for both English-only and Few-Shot scenarios. 
Overall, we find that a low resource setting with 5 human-annotated
examples in the target language ($\mathcal{D}_{l,5}$) is useful
for scaling QA to multiple languages.  More specifically, 5-shot
prompt tuning gives an EM improvement of 11.7\% absolute (58.5\% $\rightarrow$ 70.2\%) in exact match
answer accuracy on the \tydiqagoldp{} evaluation set over mT5
fine-tuned on English data only (Baseline), 3.7\% (66.5\% $\rightarrow$ 70.2\%) over mT5 fine-tuned
on~5 examples per language (Few-shot Baseline), and~4.1\% (66.1\% $\rightarrow$ 70.2\%) over mT5
fine-tuned on the data obtained with the MT approach.

\textsc{QAmeleon} further improves over the few-shot results obtained
by prompting code-davinci-002 \cite{chen2021evaluating}, PaLM-540B \cite{chowdhery2022palm}, and Flan-U-PaLM-540B \cite{chung2022scaling}, with a  similar number of available labeled examples. These approaches directly employ  extremely large PLMs for the task of QA, whereas \textsc{QAmeleon}  leverages data synthesis to distill a PLM into a much smaller mT5-XL model.
It also is important to note that \textsc{QAmeleon} as an approach is orthogonal and possibly complementary to  any improvements due to  more performant QA models and more sophisticated PLMs (e.g.,~Flan-U-PaLM-540B). 

In both English-only and Few-shot resource scenarios,
\textsc{QAmeleon} outperforms the other two data generation approaches, Machine Translation (MT), and Prompt Engineering
(PE). Despite employing PE in two stages, chain-of-thought style, we
observe that the generated data leads to lower QA
performance. Moreover, we see better performance with using
English-Only data in comparison to the Few-Shot scenario suggesting
that the PLM is able to better utilize high-quality English data
rather than small amounts of labeled data (in other languages). Finally, augmenting PLM generated data (either via PE or PT) with data
generated via MT leads to gains in QA performance over using any of
these methods independently. This could be due to the coupling of
diverse QA data i.e.,~language-specific content and  task-specific
English-centric translated content.

\begin{table}[t]
    \centering
    \begin{tabular}{l|cc} 
       \textbf{Method}      & \textbf{n-Shot} & \textbf{Avg EM} \\
       \hline \hline
       Baseline     &      ~~1 &       63.7    \\
       \textsc{QAmeleon} (PT)          &     ~~1 &       69.7     \\
       \hline
       Baseline     &      ~~5 &   66.5         \\
       \textsc{QAmeleon} (PT)          &     ~~5 &   70.2         \\
       \hline
       Baseline     &     50 &   69.3          \\
       \textsc{QAmeleon} (PT)           &     50 &   73.7         \\
       \hline
              Baseline     &      ~~100 &       70.6     \\
       \textsc{QAmeleon} (PT)          &     ~~100 &   71.9        \\
       \hline

       Supervised   &   Multi-k &   78.4         \\ \hline \hline
    \end{tabular}
    \caption{Comparison of QA performance from fine-tuning mT5-XL on 1 to 100 examples (Baseline), on synthetic data generated with prompt tuning (PT), or on the full \tydiqagoldp{} training set (Supervised). Results are averaged across languages.
    }
    \label{tab:slr_vs_lr}
\end{table}

Table~\ref{tab:lang} shows QA performance in individual languages,
for each of the methods in Table~\ref{tab:lang} in their best
performing setting: Few-shot Baseline, Machine Translation (MT),
Prompt Engineering (PE), Prompt Tuning (PT) and augmenting PE and PT with MT. Data generated by
\textsc{QAmeleon} (PT) using 5 target examples provides the best
performance in Bengali, Finnish and Telugu. A boost can be seen
for Arabic, Indonesian, Russian and Swahili when \textsc{QAmeleon} data is combined with MT
data. Language distribution listed under `\% tokens in PLM'  reflects the extremely low representation of many languages in the pre-training corpora of the PLM used in this work. As an upper bound, we additionally show the performance of supervised mT5-XL
fine-tuned on large amounts of gold training data (see
Table~\ref{tab:data}) to illustrate the remaining gap, which could
potentially be bridged by increasing the number of labeled examples or by improved (e.g. more multilingual or FLAN-tuned) PLMs. 

\paragraph{Increasing Labeled Examples Improves QA Performance} So
far, we have tested \textsc{QAmeleon} in an extremely low resource
setting, using only 5~examples in the target language. We next examine
its performance when we vary the number
of annotated examples. Table~\ref{tab:slr_vs_lr} compares the
performance of mT5-XL fine-tuned on 1 to 100 examples (Baseline), on
synthetic QA datasets generated by \textsc{QAmeleon} using corresponding number of
examples, and as an upper bound on the entire \tydiqagoldp{}
dataset. As can be seen, increasing the number of examples from 1 to
50 improves the performance of QA models. We observe however a decrease in
performance at 100 examples, showing that further research will likely be needed
to close the gap between our method and the fully supervised upper bound, while
still only labeling a small number of examples. It is
important to note that for all the amounts of available 
annotated data we considered, significant improvements in multilingual QA
could be obtained
by generating data with \textsc{QAmeleon} instead of fine-tuning the
QA model directly on labeled data.

\paragraph{The Larger the Synthetic Data, the Better the QA Model}

\begin{figure}[t]
\centering
\vspace*{-.5cm}
\hspace*{-.25cm}
\includegraphics[width=1.12\columnwidth]{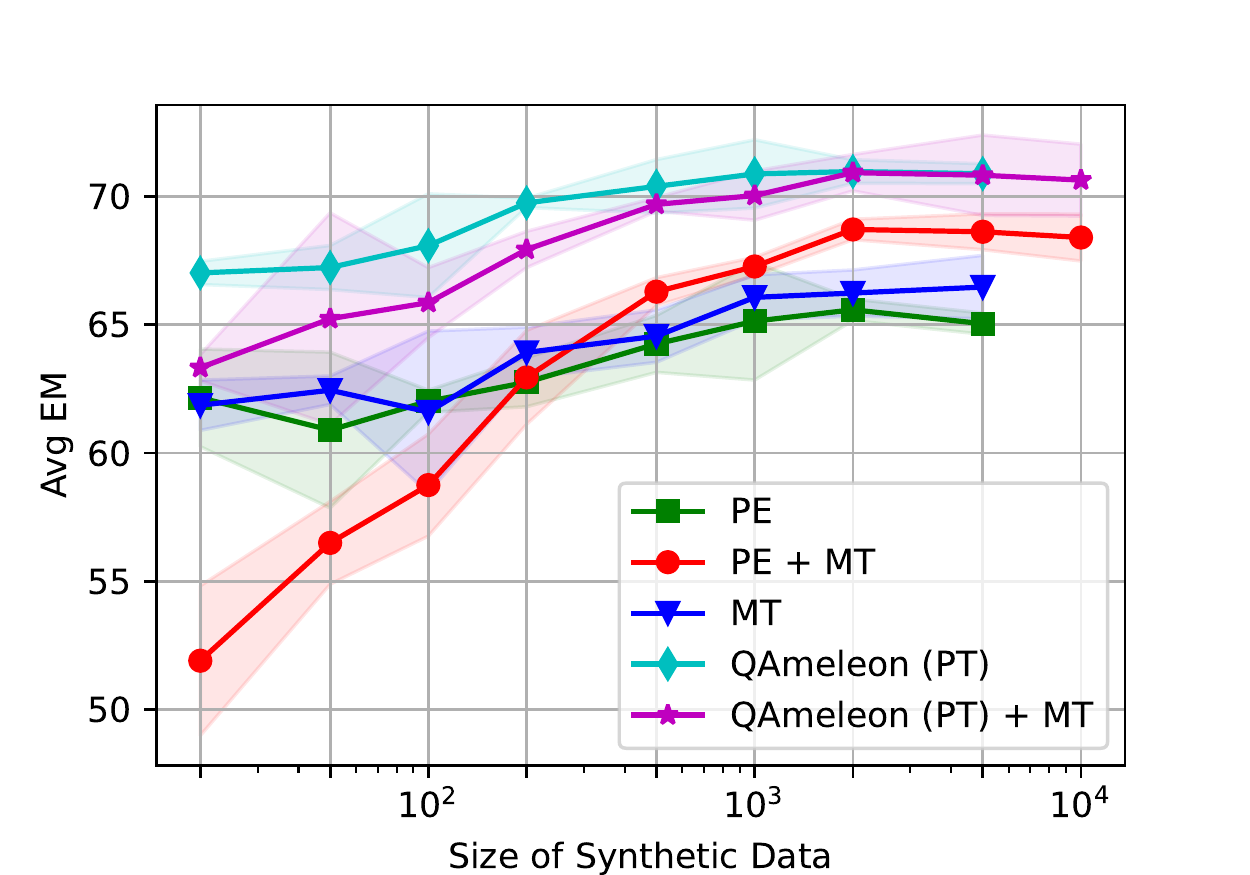}
\centering
\vspace*{-.6cm}
\caption{Effect of synthetic data size on downstream QA performance (Average EM on \tydiqagoldp{} evaluation set); results shown for mT5-XL QA model
  fine-tuned via Machine Translation (MT), Prompt Engineering (PE), 
  Prompt Tuning (\textsc{QAmeleon} (PT)),  and combinations thereof (PE + MT and \textsc{QAmeleon} (PT) + MT).
}
\label{fig:saturation_pe}
\end{figure}

We now study the impact of varying the size of the automatically
generated datasets on QA performance. As shown in
Figure~\ref{fig:saturation_pe}, when larger amounts of synthetic data
are used for fine-tuning the QA model, absolute accuracy increases. This
upshot is higher when combining PLM-generated data with Translation
data in comparison to individual datasets. This can be explained due
to the increased diversity of the combined data, which include English-centric translated
content and target language-specific content obtained from the
PLM. Eventually, we observe a saturation effect, i.e.,~beyond $O(1,000)$
QA pairs in the target language improvements are limited.

\begin{table}[t]
    \centering \resizebox{\columnwidth}{!}{
    \begin{tabular}{l|ccc}  
      \multicolumn{1}{c|}{\textbf{Method}}  &  \textbf{n-Shot} & \textbf{Avg BLEU}  &  \textbf{Avg EM} \\
       \hline \hline
       mT5-XL              &      5 &	24.74 &	57.3   \\
       \textsc{QAmeleon} (PT)   &    5 & 24.29  & 70.2 \\ \hline \hline
    \end{tabular}}
    \caption{BLEU scores and downstream QA performance on
      \tydiqagoldp{} for questions generated by mT5-XL and
      \textsc{QAmeleon} (Few-shot setting, 5 examples in
      the target language).
      }
    \label{tab:bleu_vs_qa}
\end{table}

\paragraph{BLEU Does not Correlate with  Downstream QA
  Performance} An interesting question is whether improvements in QA
performance are due to better (e.g., more grammatical or
diverse) questions. We assessed the quality of questions generated by
\textsc{QAmeleon} (PT) on \tydiqagoldp{} by measuring their
similarity to gold standard questions. We compare this with an mT5-XL model for question generation fine-tuned in a Few-shot setting.  Both \textsc{QAmeleon} (PT) and mT5-XL question generation models were given the same number of examples in each language. Table~\ref{tab:bleu_vs_qa}
reports BLEU \cite{papineni2002bleu} scores for these two models;  we additionally report  question answering performance (in terms of EM) via another set of mT5-XL models fine-tuned on the synthetic data generated by the respective models.

Even
though \mbox{mT5-XL} produces questions with slightly higher BLEU score,
\textsc{QAmeleon} generates QA data that leads to much higher QA
performance.  The result underscores the need for better
trustworthy automatic evaluation metrics \cite{sellam2020bleurt}
across languages.

\begin{table}[t]
    \centering
    \resizebox{\columnwidth}{!}{
    \begin{tabular}{l|ccc} 
       \multicolumn{1}{c|}{\textbf{Method}}     & \textbf{Avg EM} & \textbf{Avg F1}\\ \hline
       \hline
       % See https://colab.corp.google.com/drive/1GOviyKFbdKokoZzKsOIxh6B6cFhrgt1y#scrollTo=8j1FSqJnX541
       % These are dev results.
       %   English-Only   & 54.7  & 72.1 \\
       %   MT         &  58.0  & 75.5 \\
       %   \textsc{QAmeleon} (PT) & 57.3 & 74.7 \\
       %   \textsc{QAmeleon} (PT) + MT & \textbf{58.8} & 75.9 \\
       % These are test results.
       English-Only   &  53.1 & 71.8 \\
       MT             &  56.4 & 74.8 \\
       \textsc{QAmeleon} (PT) & 55.0 & 74.3 \\
       \textsc{QAmeleon} (PT) + MT & \textbf{56.8} & \textbf{75.3} \\
       \hline
       mT5-XL \cite{xue2021mt5} & 54.5 & 73.5 \\
       XLM-E-XL \cite{chi2022xlm} &  57.9 & 76.2
       \\ \hline \hline
    \end{tabular}}
    \caption{Downstream QA performance on the \textsc{MLQA} test set with
       an mT5-XL model trained on \textsc{SQuAD} English data
      (English-Only),  \textsc{SQuAD} translated to all 
      \textsc{MLQA} languages (MT), on synthetic data generated by
      \textsc{QAmeleon} (5-shot) in all  \textsc{MLQA} languages,
      or on a combination of data generated by MT and 
      \textsc{QAmeleon}. Results for \newcite{xue2021mt5} and
      \newcite{chi2022xlm} are taken from the respective papers. 
    }
    \label{tab:mlqa}
\end{table}

\paragraph{Our Results Generalize to MLQA}
To validate the general
applicability of our approach, we evaluate \textsc{QAmeleon} on MLQA
\cite{lewis-etal-2020-mlqa}. We prompt-tune the PLM on 5 examples per
language taken from the MLQA development set, since MLQA does not
provide training partitions. We generate synthetic datasets in all of
the MLQA languages and compare an English-only baseline, MT,
and \textsc{QAmeleon} (PT) approaches as we did previously for
\tydiqagoldp{}. We report results (EM and F1) using \mbox{mT5-XL} as the QA model
in Table~\ref{tab:mlqa}, where English is included in the average
performance.

We find that the MT approach is very
effective on MLQA, which is not surprising since MLQA questions are
translated from English. \textsc{QAmeleon} (PT), however, still delivers
an improvement in combination with MT synthetic data.
Table~\ref{tab:mlqa} further  reports comparisons with the state-of-the-art models of \newcite{xue2021mt5} and
      \newcite{chi2022xlm}. The former is mT5-XL (3.7B~parameters) fine-tuned on English data only, whereas XLM-E-XL (2.2B parameters)  benefits from 
      a different language model pretraining technique.  The latter approach is orthogonal and potentially complementary to \textsc{QAmeleon}.

\begin{table}[t]
    \centering
    \resizebox{\columnwidth}{!}{
    \begin{tabular}{l|cc|c} 
                   & \multicolumn{2}{c|}{\textbf{\tydiqagoldp{}}} & \textbf{MLQA} \\
         \textbf{Language}  & PE & \textsc{QAmeleon} & \textsc{QAmeleon}  \\
         
       \hline \hline
         Arabic & 5,219 & 8,499 & 14,738 \\
         Bengali & 5,948 & 8,036 & ---\\
         Chinese & --- & --- & 14,669 \\
         Finnish & 8,062 & 5,943 & --- \\
         German & --- & --- & 11,186 \\
         Hindi &--- & --- & 12,036 \\
         Indonesian & 6,487 & 7,810 & --- \\
         Kiswahili & 8,003 & 8,041 & --- \\
         Korean & 5,229 & 7,906 & --- \\
         Russian & 5,619 & 7,441 & --- \\
         Spanish & --- & --- & 10,134 \\
         Telugu & 2,742 & 5,222 & ---  \\
         Vietnamese & --- & --- & 13,333 \\
       \hline 
         Total & 47,309 & 52,955 & 89,344 \\ \hline \hline
    \end{tabular}}
    \caption{Number of synthetic question-answer pairs per language
      generated via Prompt Engineering (PE) and \textsc{QAmeleon} (PT)
      with  5 human-labeled examples.
    % \textcolor{red}{todo: add MLQA stats, and can be merged with table1}
    }
    \label{tab:syn_data}
\end{table}

\section{Data Analysis}
\label{sec:data-analysis}

\begin{table*}[t]
    \centering
    \resizebox{1.0\textwidth}{!}{
    \begin{tabular}{c|p{3.7in}|p{3.7in}} 
        Lang& \multicolumn{1}{c}{Human Annotated} &
        \multicolumn{1}{|c}{\textsc{QAmeleon} (PT)}  \\
      \hline\hline
         \textbf{ar} &  \textbf{Q:}
         \begin{otherlanguage}{arabic}
         \textRL{متى تأسست جامعة فرايبورغ؟}
         \end{otherlanguage} \hfill\textbf{ A:} 1457 &	\textbf{Q:} 
         \begin{otherlanguage}{arabic}
         \textRL{متى تأسست جامعة فرايبورغ؟}
         \end{otherlanguage} \hfill\textbf{ A:} 1457 \\
          & \textit{Q: When was the University of Freiburg established?} \hfill\textit{A: 1457} & \textit{Q: When was the University of Freiburg founded?} \hfill\textit{A: 1457} \\
           \textbf{sw} &
           \textbf{Q:} Je, Kifaru ana urefu kiasi gani? \hfill\textbf{A:} "mita 3.5-4.6 & 
            \textbf{Q:} Je , faru mweupe ana uzito gani? \hfill\textbf{A:} kilogramu 3,500 \\
           & \textit{Q: How tall is a Rhino?} \hfill\textit{A: 3.5-4.6 meters}  & \textit{Q: How much does a white rhino weigh?} \hfill\textit{A: 3,500 kilograms}\\
           \textbf{ru} & 
           \textbf{Q:} \begin{otherlanguage}{russian}
           Когда был подписан Георгиевский трактат? 
           \end{otherlanguage}\hspace{2cm}
           \textbf{A:} \begin{otherlanguage}{russian} 24 июля (4 августа) 1783 года\end{otherlanguage}
           & \textbf{Q:}
           \begin{otherlanguage}{russian}
           В какой крепости был заключен Георгиевский трактат?
            \end{otherlanguage}\hspace{.05cm}\textbf{A:}
           \begin{otherlanguage}{russian}
            Георгиевск (Северный Кавказ)
           \end{otherlanguage}
           \\
          & 
           \textit{Q: When was the Treaty of Georgievsk signed?} & \textit{Q: In what fortress was the Treaty of St. George concluded?} \\
           &
           \textit{A: July 24 (August 4), 1783}  & 
           \textit{A: Georgievsk (North Caucasus)}  \\
    \hline \hline
    \end{tabular}}
    \caption{Examples of QA pairs from human-annotated \tydiqa{}
      and generated by \textsc{QAmeleon} (PT) on corresponding
      passages. English translations from Google Translate are added for
      readability.}
    \label{tab:examples}
\end{table*}

 Table~\ref{tab:syn_data} shows the size of synthetic data resources
 generated via Prompt Engineering (PE) and \textsc{QAmeleon} (PT), per
 language and in total. These were in the range of
 47,000-53,000 QA examples for \tydiqagoldp{}, and 89,000 for
 MLQA. The varying size of the data across languages is due to the
 filtering described in Section~\ref{sec:exp}. In some languages
 (e.g., Telugu) generation is more noisy leading to fewer
 data points. We conjecture this is due to the PLM being exposed to
 less data representing these languages during pre-training. We
 further hypothesize that a more multilingual pre-training of PLMs
 could potentially lead to better quality data across all languages.

 Machine translation (MT) creates the same number of data points as
 the source training set. For \tydiqagoldp{}, the English training
 contains~3,696 data points (Table \ref{tab:data}), leading to
 approximately 29,000 QA examples across 8 languages. In MLQA, machine
 translation (MT) uses SQuAD as the English dataset, consisting of
 $\sim$ 87,000 data points, leading to $\sim$ 522,000 QA examples
 across~6 languages.

\begin{figure*}[t]
    \centering
    \begin{subfigure}[b]{0.3\textwidth}
         \centering
         \includegraphics[width=\textwidth]{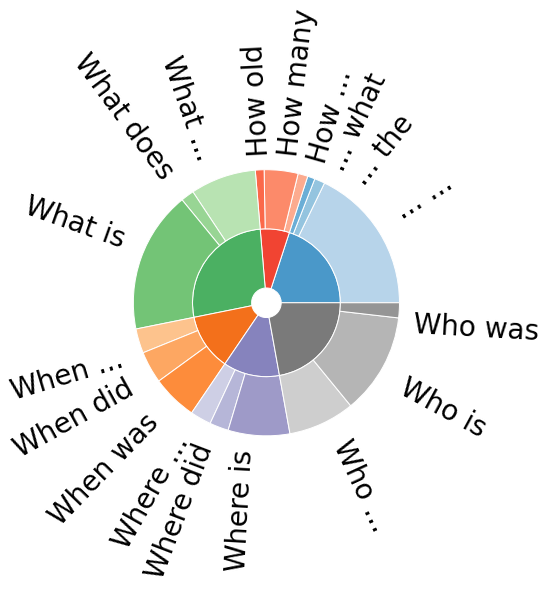}
         \caption{Indonesian: \textsc{QAmeleon} (PT)}
         \label{fig:fi_chart}
     \end{subfigure}
     \hfill
     \begin{subfigure}[b]{0.3\textwidth}
         \centering
         \includegraphics[width=\textwidth]{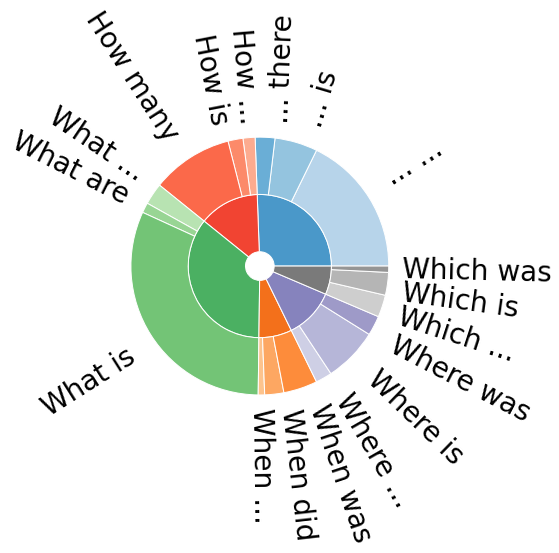}
         \caption{Telugu: \textsc{QAmeleon} (PT)}
         \label{fig:te_chart}
     \end{subfigure}
     \hfill
     \begin{subfigure}[b]{0.3\textwidth}
         \centering
         \includegraphics[width=\textwidth]{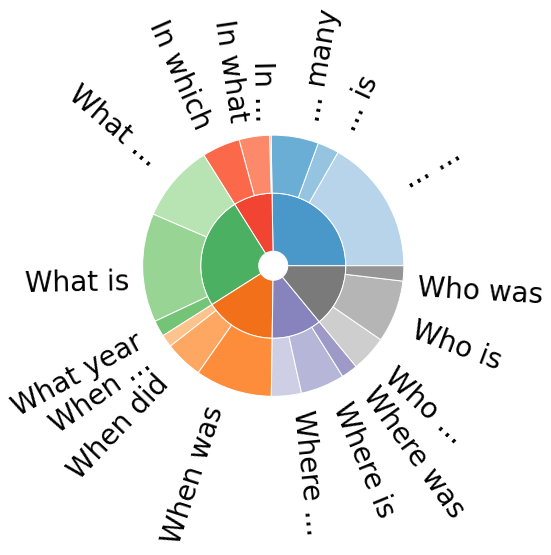}
         \caption{All: \textsc{QAmeleon} (PT)}
         \label{fig:te_chart}
     \end{subfigure}
     \hfill
     \begin{subfigure}[b]{0.3\textwidth}
         \centering
         \includegraphics[width=\textwidth]{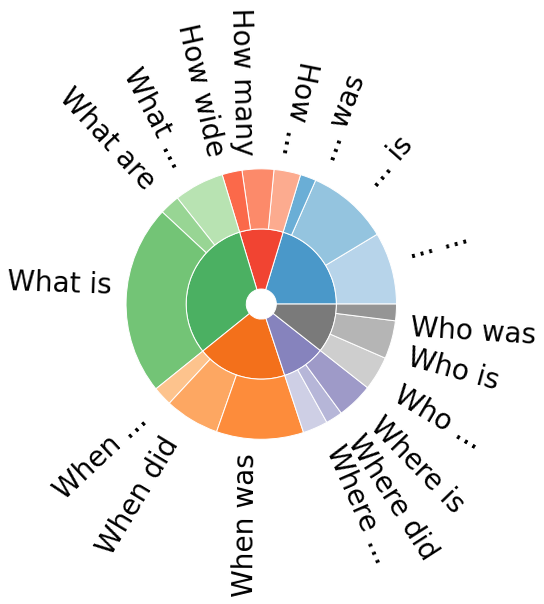}
         \caption{Indonesian: \tydiqagoldp{}}
         \label{fig:te_chart}
     \end{subfigure}
     \hfill
     \begin{subfigure}[b]{0.3\textwidth}
         \centering
         \includegraphics[width=\textwidth]{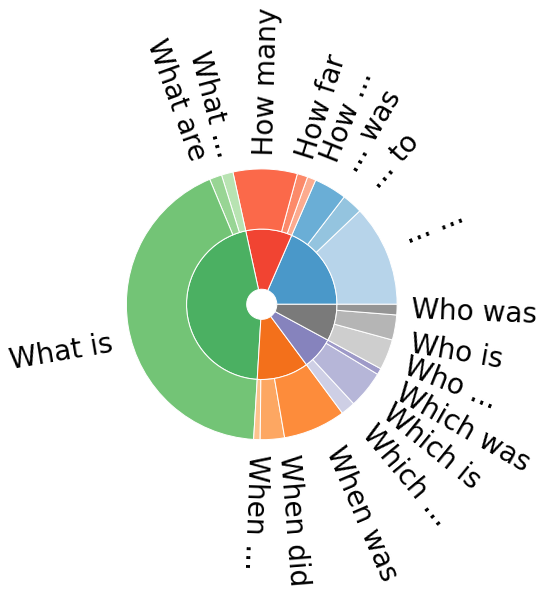}
         \caption{Telugu: \tydiqagoldp{}}
         \label{fig:te_chart}
     \end{subfigure}
     \hfill
     \begin{subfigure}[b]{0.3\textwidth}
         \centering
         \includegraphics[width=\textwidth]{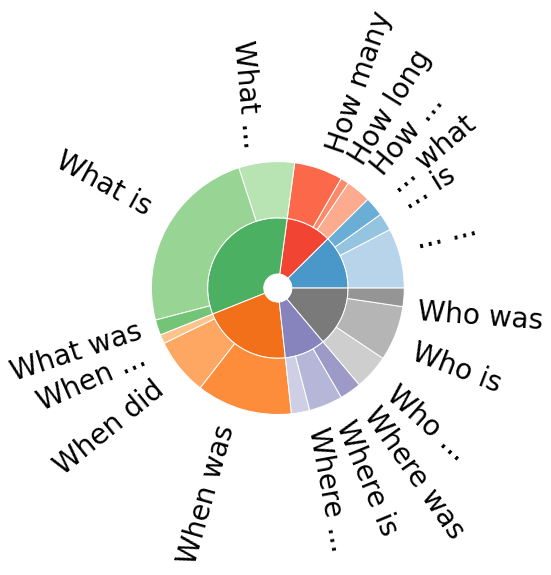}
         \caption{All: \tydiqagoldp{}}
         \label{fig:te_chart}
     \end{subfigure}
     \caption{Distribution of question category for \textsc{QAmeleon} (PT) generated questions (a,b,c) and \tydiqagoldp{} training questions (d,e,f). Category are obtained by translating the questions to English with Google Translate and grouping by the first two word tokens.}
    \label{fig:question_analysis}
\end{figure*}

\begin{table*}[t]
    \centering
    \resizebox{0.9\textwidth}{!}{
    \begin{tabular}{c|p{5.5in}} 
         Language  & \multicolumn{1}{c}{
          \textsc{QAmeleon} (PT)} \\
      \hline \hline
      \textbf{fi} & \textbf{Q:}
      \begin{otherlanguage}{finnish}
      Milloin Tullintori valmistui? 
      \end{otherlanguage}\hfill
      \textbf{A: }\textit{1990}
      \\
      & \textit{Q: When was Tullintori completed?} \hfill\textit{A: 1990}  \\
     \textbf{ru} & \textbf{Q:} \begin{otherlanguage}{russian}
      Какой отраслью экономики преимущественно занято население Узбекистана?
     \end{otherlanguage}
     \textbf{A:} \begin{otherlanguage}{russian}
     сфера обслуживания и туризма
     \end{otherlanguage}
      \\
          & \textit{Q: What sector of the economy is predominantly employed by the population of Uzbekistan? A: service and tourism}
          \\
    \textbf{sw} & \textbf{Q: }Nyoka birisi iko katika familia gani? \hfill\textbf{A:} Typhlopidae \\
    & \textit{Q: Which family is the Birsi snake in?} \hfill\textit{A: Typhlopidae} \\
    \textbf{ar} &\textbf{Q: }
    \begin{otherlanguage}{arabic}
         \textRL{
     في أي ولاية تقع بلدة فرانكلين ، مقاطعة مانيتووك؟
     } \end{otherlanguage}
     \hfill\textbf{A: }\begin{otherlanguage}{arabic}\textRL{ويسكونسن} \end{otherlanguage}\\
    & \textit{Q: In which state is Franklin Township, Manitowoc County?} \hfill\textit{A: Wisconsin} \\
    \textbf{id}&\textbf{Q:} Siapa pencipta manga DN Angel? \hfill\textbf{A:} Yukiru Sugisaki \\
    & \textit{Q: Who created the manga DN Angel?} \hfill\textit{A: Yukiru Sugisaki}    \\
    \hline 
    \hline 
    \end{tabular}}
    \caption{QA pairs (random selection) generated by
      \textsc{QAmeleon} (PT) on Wikipedia passages. English
      translations from Google Translate are added for readability.}
    \label{tab:examples_qa}
\end{table*}

Figure~\ref{fig:question_analysis} shows the distribution of various question types for individual languages and on average across all languages.  For each language, synthetically generated questions were first translated to English and then grouped into categories (inner circle) based on their first word and  sub-categories (outer circle) based on the first two words. We find that \textsc{QAmeleon} (Figure~\ref{fig:question_analysis}~(c)) generates more diverse questions in comparison to \tydiqagoldp{} (Figure~\ref{fig:question_analysis}~(f)). The distribution of question words varies across languages in both the datasets. For example, diversity is higher for Russian, Finnish, and Indonesian (Figure~\ref{fig:question_analysis}~(a,d)); however, for Bengali and  Telugu (Figure~\ref{fig:question_analysis}~(b,e)), the distribution of questions is skewed towards a specific question type (`What' for Telugu and `Other' for Bengali). This could be attributed to a lack of diversity in questions for these languages or poor translation quality leading to skewed utterances. 

Table~\ref{tab:examples} illustrates
randomly sampled examples of QA pairs generated by \textsc{QAmeleon}
(PT) for passages in the \tydiqagoldp{} eval set. For these
passages, we also have access to  human annotated QA pairs. As
can be seen, QA pairs generated by \textsc{QAmeleon} are of similar
quality and at times more diverse compared to the human-annotated dataset. Table~\ref{tab:examples_qa} illustrates
examples of QA pairs generated by \textsc{QAmeleon} from randomly selected Wikipedia passages. 

\section{Related Work}

\paragraph{Data Generation for QA} Prior work on the generation of QA data
has mostly focused on English and typically divides the task into
answer extraction/generation and question generation, followed by some
type of filtering. \citet{alberti-etal-2019-synthetic} employ
round-trip consistency for filtering with BERT-based models. Other
work \cite{shakeri-etal-2020-end} uses BART to jointly generate a
question and its answer given an input passage, employing
likelihood-based filtering. \citet{lewis-etal-2021-paq} use a
RoBERTa-based passage selection model to identify interesting
passages. \citet{bartolo-etal-2021-improving} additionally train the
generation models on an adversarial QA dataset, while
\citet{yao-etal-2022-ais} integrate a QA-pair ranking module.

The above approaches generally require large amounts of labeled QA
data in the form of \textsc{SQuAD} \cite{rajpurkar-etal-2016-squad} or
Natural Questions \cite{kwiatkowski-etal-2019-natural} to train
passage selection and question generation models. In contrast, we only
assume access to a few question-answer pairs per language.

\paragraph{Multilingual QA} In this work we used mT5-XL \cite{xue2021mt5} as our
 reference QA model.
 We note that a slightly more performant choice could have been ByT5 \cite{xue2022byt5},
 which reports improvements on \tydiqagoldp{} by operating directly on raw text instead
 of sentence pieces.
 Existing work on low resource multilingual QA 
 has been relatively limited. \citet{lee-etal-2018-semi} propose to
 use automatically translated high-confidence QA examples for
 training, while other approaches
 \cite{kumar-etal-2019-cross,chi2020cross} only generate questions and
 require supervised training data in the target language. Other approaches
 \cite{riabi-etal-2021-synthetic,shakeri-etal-2021-towards,kramchaninova-defauw-2022-synthetic}
 focus on zero-shot transfer, i.e.,~a multilingual model trained on
 QA data generation on \textsc{SQuAD} (and optionally automatically
 translated \textsc{SQuAD} data) is applied to other languages. Our
 work shows that few-shot settings result in better multilingual
 generation quality in comparison to zero-shot models.

\paragraph{Prompting} Existing work  \citep[\emph{inter
    alia}]{brown2020language,schick-schutze-2021-just} has shown that
prompting pre-trained large language models can lead to strong
performance in a wide range of tasks including natural language
generation and common sense reasoning. In the context of multilingual
QA, \citet{chowdhery2022palm} employ a single prompt and a few labeled
examples in the target language. In contrast, we employ chain-of-thought prompting, and English answers and questions as a bridge. Moreover, our experiments
with \textsc{QAmeleon} demonstrate that prompt tuning is superior and
a viable alternative to large-scale annotation. Prompting in multilingual settings has achieved the best performance using English prompts and target language exemplars \cite{winata-etal-2021-language,Lin2022,Shi:ea:2022}. We demonstrate that parameter-efficient methods such as prompt tuning using target language exemplars \cite{lester-etal-2021-power} is a superior choice.

\label{sec:bibtex}

\section{Benefits and Limitations}

The method proposed in this work, QAmeleon,  prompt tunes large PLMs to generate multilingual synthetic question answering data. In this section we discuss its benefits over related approaches, but also drawbacks and limitations. The main benefits are  large performance improvements over alternative methods, as borne out by our experiments, as well as surprising data efficiency achieved through  large-scale  pre-training and a few manual annotations. Alternative methods considered here are multilingual QA approaches for low resource languages, such as translate-test, translate-train, fine-tuning multilingual models directly on the small amount of available training data, performing multilingual QA directly through in-context learning, or even synthetic data generation with prompt engineered PLMs.

Another benefit of our approach stems from prompt tuning itself, which is able to learn from a tiny number of training examples, as low as one example per language in our experiments, whereas fine-tuning cannot be utilized as easily. Prompt tuning also affords the practical advantage of being space efficient; a  fraction of a percent of the storage space is used to save the learned parameters, since only the learned soft prompt needs to be stored. Our evaluation methodology also provides benefits, since we measure question answer performance directly on downstream models instead of using a proxy like BLEU or ROUGE on generated questions. As shown in Table~\ref{tab:bleu_vs_qa} proxy metrics can be misleading, and one might conclude that smaller models generate better questions than large PLMs if the evaluation were to consider only question BLEU scores.

The main drawback of our method is the high computational cost to prompt tune the PLM and to generate the synthetic data. While prompt tuning is not as expensive as fine-tuning, we still need to perform optimization on a model containing hundreds of billions of parameters. We estimate the cost of each prompt tuning and data generation experiment to be in the order of 256 TPU v3 chips for 12 hours. Another limitation of our experimental results is that they are fundamentally tied to a specific large PLM. PLMs are an area of  active research, so any improvements in  pre-training techniques, construction of pre-training sets, instruction tuning or reinforcement learning, are likely to translate in improvements for our synthetic data generation method. Promising areas of future work are parameter efficient techniques similar to prompt tuning, as well as analysis of data augmentations techniques like QAmeleon across different types and sizes of PLMs. Moreover, a more formal understanding of how the number of manual annotations (aka few shots) interacts with the quality of synthetic generation, would also be useful. Perhaps somewhat counter-intuitively, our experiments showed that QA performance does not drastically improve when scaling from 50~to~100 manual examples. 

\section{Data Release}

To assist with  the replicability of our results and to allow other researchers to benefit from our work, we will release a significant portion of the synthetic data generated by \textsc{QAmeleon} in the 5-shot scenario. To minimize the chance that question-answer pairs generated by the PLM contain sensitive, offensive or controversial material, we vetted each generated question with three human raters. We asked each rater to discard  question-answer pairs that made generalized claims about groups, contained opinions that were potentially disparaging or embarrassing to one or more people, or names of individuals not related to media (e.g., film, TV) or sport. The release will contain 47,173 examples, each with a Wikipedia passage, a question and an extractive answer, corresponding to 89\% of the examples utilized in this work for the 5-shot scenario.

\section{Conclusions}

In this work, we examined the ability of pre-trained language models
to generate synthetic data for bootstrapping multilingual QA systems,
with as few as five examples in a new target language. We introduced
\textsc{QAmeleon}, a parameter efficient approach which uses prompt
tuning to automatically create multilingual QA data. Extensive
experiments under different resource scenarios demonstrate that
\textsc{QAmeleon} is superior to prompt engineering and competitive
baselines based on machine translation. In the future, we would like
to extend this approach to other multilingual tasks, including retrieval,
summarization, and semantic parsing.

\bibliography{tacl2021,anthology,custom}
\bibliographystyle{acl_natbib}

\end{document}